\documentclass[10pt,twocolumn,letterpaper]{article}

\usepackage{iccv}
\usepackage{times}
\usepackage{epsfig}
\usepackage{graphicx}
\usepackage{amsmath}
\usepackage{amssymb}
\usepackage{booktabs}
\usepackage{multirow}
\usepackage[table,x11names, svgnames, dvipsnames]{xcolor}
\usepackage{adjustbox}

\usepackage[pagebackref=true,breaklinks=true,letterpaper=true,colorlinks,bookmarks=false]{hyperref}
\usepackage[capitalize]{cleveref}
\crefname{section}{Sec.}{Secs.}
\Crefname{section}{Section}{Sections}
\Crefname{table}{Table}{Tables}
\crefname{table}{Tab.}{Tabs.}

\iccvfinalcopy % *** Uncomment this line for the final submission

\def\iccvPaperID{10418} % *** Enter the ICCV Paper ID here

\ificcvfinal\fi

\begin{document}

%%%%%%%%% TITLE
\title{Audio-Visual Deception Detection: DOLOS Dataset and Parameter-Efficient Crossmodal Learning}

\author{Xiaobao Guo$^{12}$\thanks{Equal contribution} \and Nithish Muthuchamy Selvaraj$^3$\footnotemark[1] \and Zitong Yu$^3$\thanks{Corresponding author} \and Adams Wai-Kin Kong$^2$\and Bingquan Shen$^4$ \quad \quad Alex Kot$^3$\\
\normalsize$^1$Rapid-Rich Object Search (ROSE) Lab, Interdisciplinary Graduate Programme, Nanyang Technological University \\
\normalsize$^2$School of Computer Science and Engineering, NTU \quad \normalsize$^3$School of Electrical \& Electronic Engineering, NTU\\
\normalsize$^4$DSO National Laboratories, Singapore\\
% {\tt\small xiaobao001@e.ntu.edu.sg}\ {\tt\small \{ms.nithish, zitong.yu, adamskong, eackot\}@ntu.edu.sg} {\tt\small sbingqua@dso.org.sg}
{\tt\small\{xiaobao001, ms.nithish, zitong.yu, adamskong, eackot\}@ntu.edu.sg sbingqua@dso.org.sg}
}
\vspace{-1.5em}
% %%%%%%%%% TITLE
% \title{\LaTeX\ Author Guidelines for ICCV Proceedings}

% \author{Xiaobao Guo$^1$\thanks{Equal contribution} \and Nithish Muthuchamy Selvaraj$^1$\footnotemark[1] \and Zitong Yu$^1$\thanks{Corresponding author} \and Adams Kong$^1$ \and Bingquan Shen$^2$ \\
% $^1$Nanyang Technological University, Singapore\\
% $^1$Rapid-Rich Object Search (ROSE) Lab, IGP, Nanyang Technological University, Singapore \\
% % {\tt\small xiaobao001@e.ntu.edu.sg}
% % % For a paper whose authors are all at the same institution,
% % % omit the following lines up until the closing ``}''.
% % % Additional authors and addresses can be added with ``\and'',
% % % just like the second author.
% % % To save space, use either the email address or home page, not both
% \and
% Alex Kot$^1$\\
% $^2$DSO National Laboratories, Singapore
% % {\tt\small sbingqua@dso.org.sg}
% }

\maketitle
\ificcvfinal\thispagestyle{empty}\fi

%%%%%%%%% ABSTRACT
\begin{abstract}
% Research on predicting human behaviors continues to grow in importance with the development of human-centered AI. 
Deception detection in conversations is a challenging yet important task, having pivotal applications in many fields such as credibility assessment in business, multimedia anti-frauds, and custom security. Despite this, deception detection research is hindered by the lack of high-quality deception datasets, as well as the difficulties of learning multimodal features effectively. To address this issue, we introduce DOLOS\footnote {The name ``DOLOS" comes from Greek mythology.}, the largest gameshow deception detection dataset with rich deceptive conversations. DOLOS includes 1,675 video clips featuring 213 subjects, and it has been labeled with audio-visual feature annotations. We provide train-test, duration, and gender protocols to investigate the impact of different factors. We benchmark our dataset on previously proposed deception detection approaches. To further improve the performance by fine-tuning fewer parameters, we propose Parameter-Efficient Crossmodal Learning (PECL), where a Uniform Temporal Adapter (UT-Adapter) explores temporal attention in transformer-based architectures, and a crossmodal fusion module, Plug-in Audio-Visual Fusion (PAVF), combines crossmodal information from audio-visual features. Based on the rich fine-grained audio-visual annotations on DOLOS, we also exploit multi-task learning to enhance performance by concurrently predicting deception and audio-visual features. Experimental results demonstrate the desired quality of the DOLOS dataset and the effectiveness of the PECL. The DOLOS dataset and the source codes are available at~\href{https://github.com/NMS05/Audio-Visual-Deception-Detection-DOLOS-Dataset-and-Parameter-Efficient-Crossmodal-Learning/tree/main}{here}.

\end{abstract}

%%%%%%%%% BODY TEXT
\section{Introduction}

Deception is a pervasive and complex phenomenon that occurs in all areas of life, and understanding its nature and impact is crucial in preventing negative consequences.
Effective deception detection has crucial implications in the area of border security, anti-fraud, business negotiations and etc.~\cite{graciarena2006combining,fornaciari2013automatic,abouelenien2016detecting,gogate2017deep,ding2019face,kong2022beyond,stathopoulos2020deception}. Deep learning algorithms have achieved comparable or even better performance than human in many complex tasks~\cite{alphafold,alphazero,go,gpt-3,clip}. One may expect AI models can also bring significant breakthroughs in deception detection. Albeit the fruitful progress in computer vision~\cite{vaswani2017attention,dosovitskiy2020vit, liu2021swin,tolstikhin2021mlp} and audio representation learning~\cite{baevski2020wav2vec,hsu2021hubert,kong2020panns}, it remains a significant challenge to efficiently explore AI ability to process multimodal information in perceiving and predicting human deceptive behaviors.

The performance of AI models in deception detection is heavily reliant on the availability of authentic and effective deception samples from the real world. The deceptive subjects must be spontaneous and motivated~\cite{depaulo2003cues, vrij2012eliciting} such that certain behavioral cues (e.g., vocal pitch and chin raise) could be more pronounced. Public benchmark datasets collected from court trials~\cite{perez2015deception}, game shows~\cite{soldner2019box}, and lab-based scenarios~\cite{gupta2019bag}, have contributed to spurring interest and progress in deception detection research. Deception can be affected by a range of factors in diverse real-world situations. Thus, it is necessary to create deception datasets from different scenarios. However, current datasets are still insufficient to drive further progress and inspire novel ideas due to their limitations on both quantity and quality. These limitations include (1) the small number of deceptive samples and subjects, (2) the lack of rich annotated visual and speech attributes, and (3) the variety of protocols. It is imperative to build a larger and richer deception detection dataset. In particular, more deceptive samples and subjects, better annotations for facial movements, gestures, and audio attributes, and more types of protocols for investigating factors affecting deception detection.

In addition to a high-quality dataset, effective methods are equally important to deception detection. A variety of works have been done towards using visual and acoustic information in videos for deception detection~\cite{soldner2019box,gupta2019bag,perez2015deception}. Current methods fall into two categories: unimodal learning and multimodal fusion. However, both approaches have limitations as they do not fully exploit unimodal features or integrate complementary information from multiple modalities. To make better use of available information, fine-tuning large pre-trained models has shown promising results~\cite{radford2018improving, kolesnikov2020big,jia2021scaling,vaessen2022fine}. However, fully fine-tuning pre-trained models, such as W2V2~\cite{baevski2020wav2vec} and ViT~\cite{dosovitskiy2020vit}, can be inefficient and lead to overfitting, especially if the downstream dataset is limited. Therefore, it is important to consider parameter efficiency when fine-tuning pre-trained models. Adapters~\cite{houlsby2019parameter, jie2022convolutional} offer an efficient approach for fine-tuning models. While originally proposed for language~\cite{houlsby2019parameter} or vision~\cite{jie2022convolutional} tasks, adapters have not yet been applied to multiple modalities for temporal feature extraction.

\textbf{Contributions.} In this paper, we establish a new deception detection dataset and propose parameter-efficient crossmodal learning for audio-visual deception detection.

As our first contribution, we introduce DOLOS, a new gameshow dataset for audio-visual deception detection. DOLOS has several merits compared with current public datasets. First, the gameshow is a reliable source for collecting deception detection data because all the participants are motivated to cheat and the ground truths are available. Second, the proposed dataset is in a conversational setup, where the deception behaviors are more naturally presented. Third, our dataset is the largest in terms of the number of subjects and also the largest non-lab based dataset in terms of the number of video clips. The dataset has been labeled with fine-grained audio-visual annotations. We also benchmark our dataset on previous deception detection approaches that involve both unimodal and multimodal features. In addition, we offer three distinct protocols, namely train-test, duration, and gender, to explore various factors that may impact deception detection.

Our second contribution is Parameter-Efficient Crossmodal Learning (PECL), a method for deception detection that achieves high performance by fine-tuning a small number of extra learnable weights. Specifically, we introduce a Uniform Temporal Adapter (UT-Adapter) that explores the temporal attention between input embeddings for both visual and audio modalities without the need for delicate modifications. For multimodal fusion, we propose Plug-in Audio-Visual Fusion (PAVF), which utilizes the complementary information between visual and audio features to enhance the overall performance. PECL is designed to be parameter efficient, with only the UT-Adapter, PAVF modules, and the classifier being trainable. Furthermore, we explore the benefits of multi-task learning, a proven method that enhances performance in audio and visual tasks~\cite{zhang2019attention,tao2020end}. By simultaneously predicting deception, facial movements, and phonetic features, our proposed method can be further improved.

To compare and show the advantages of our dataset and the proposed method, we conducted extensive experiments. In the cross-testing with the current gameshow benchmark dataset~\cite{soldner2019box}, DOLOS performs better on several test sets in different scenarios. The experimental results on DOLOS also showed that PECL yields superior performance on different protocols. Through our experiments, we uncover valuable insights into multimodal deception detection. We believe that DOLOS, the proposed method, and the benchmarking experiments provide valuable resources to other researchers in advancing research in this area. 

\section{Related Work}

\subsection{Deception Detection Datasets} 
Early works in deception detection research investigated the psychological premises of deception \cite{depaulo2003cues,hirschberg2005distinguishing,warren2009detecting,levine2014theorizing,vrij2012eliciting} and identified potential cues, such as physiological, visual, vocal, verbal and behavioral cues. The pioneering work on multimodal deception detection emerged with the introduction of the Real-Life Trials (RLT) dataset~\cite{perez2015deception} collected from courtroom trials. The courtroom setting provides a valuable and authentic scenario of people lying, but RLT is small and its performance bottleneck was quickly reached. P{\'e}rez-Rosas~\emph{et al.}~\cite{perez2015verbal} and Kamboj~\emph{et al.}~\cite{kamboj2020multimodal} later proposed another two new datasets collected from street interviews and political speeches. However, it is difficult to verify the veracity of the responses of the subjects in these datasets. Several lab-based multimodal deception detection datasets were proposed to overcome these shortcomings. The settings of these datasets were typically in situations that speakers were allowed to lie, such as describing an object~\cite{gupta2019bag},  a personality~\cite{lloyd2019miami}, and face-to-face interviews with the subjects~\cite{speth2021deception}. However, the incentive to deceive may be low, leading to fewer deceptive cues captured.

Reality TV shows based on truth or lie games  provide an avenue to collect examples of conversational deception. They also offer a sweet spot among incentive, veracity, and sample size requirements. Often, the participants lie to each other to win the game, and the truth is revealed at the end of the round. Large amounts of gameshow footage are available online. The first gameshow-based deception detection dataset was proposed in \cite{soldner2019box} and it primarily focused on two modalities: text and manually annotated visual features. However, directly applying this dataset for multimodal deception detection by using audio and visual modalities faces several challenges, which we will describe in Sec.~\ref{sec:formatting}.

\begin{figure}[t]
\centering
\includegraphics[width=0.9\linewidth]{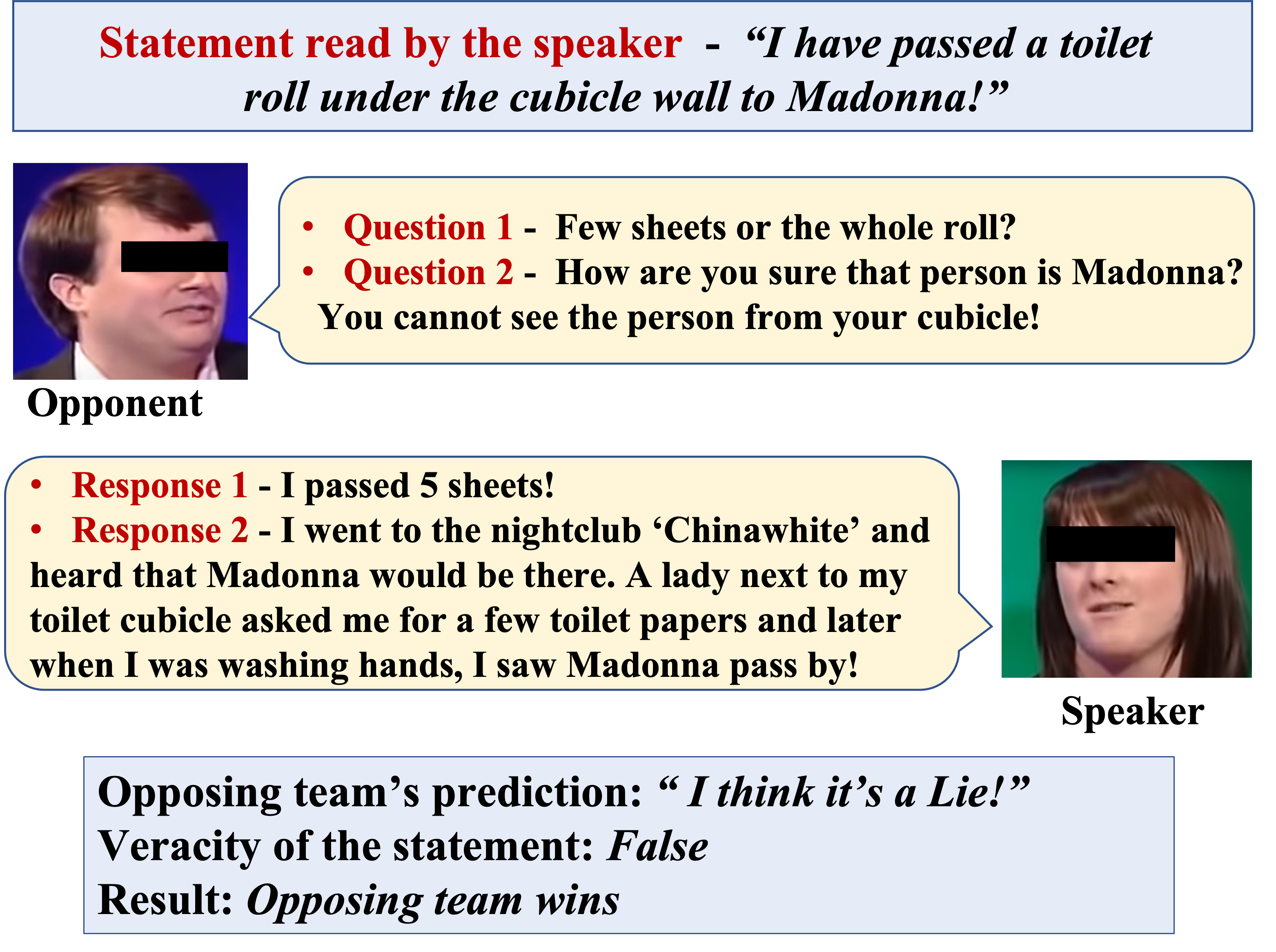}
\vspace{-0.3em}
  \caption{\small{
  Gameshow format. An example of a deceptive round.
  }
  }
\label{fig:format}
\vspace{-1.5em}
\end{figure}

\subsection{Multimodal Deception Detection}

Previous works on multimodal deception detection include using unimodal and multimodal features and exploring effective fusion mechanism~\cite{mathur2020introducing, karimi2018toward, karnati2021lienet, avola2019automatic, ding2019face, wu2018deception, gogate2017deep}. Gogate \emph{et al.}~\cite{gogate2017deep} proposed a deep model that incorporated the audio cues with visual and text modalities to improve the accuracy of deception prediction. 
Karimi \emph{et al.}~\cite{karimi2018toward} explored deceptive cues from RGB images and raw audio in an end-to-end manner. Wu \emph{et al.}~\cite{wu2018deception} utilized several types of features, including micro-expression and IDT (Improved Dense Trajectory) features from RGB images, MFCC (Mel-frequency Cepstral Coefficients) features from the audio, and transcripts. Avola \emph{et al.}~\cite{avola2019automatic} addressed the deception detection issue by directly using AUs (facial action units) from video frames and classifying them through an SVM. Ding \emph{et al.}~\cite{ding2019face} tried to combine facial expressions and body movements to extract more visual deception cues. They also tried to improve the performance by involving audio features and transcripts. Mathur and Matari\'c~\cite{mathur2020introducing} focused on facial expressions for deception detection and studied interpretable features from visual, vocal, and verbal modalities. Karnati \emph{et al.}~\cite{karnati2021lienet} proposed deep networks that learn audio, visual, and EEG representations for deception detection.  

Most of the previous works tried to enhance the performance by involving more modalities, more features, and using better feature extraction methods. However, they heavily rely on spatial features rather than temporal information within and between the modalities, which may lose potentially important cues and lead to sub-optimal results.

In terms of fusion methods, the feature-level fusion which concatenates multimodal embeddings and uses the linear layers to extract crossmodal information~\cite{karimi2018toward, avola2019automatic, ding2019face, wu2018deception, gogate2017deep} was the most popular approach. Other works 
implement decision-level fusion, such as score-level fusion~\cite{karnati2021lienet, gogate2017deep}. In all cases, performance can be boosted by a reliable and efficient fusion method.

%-------------------------------------------------------------------------

\begin{figure*}[t]
\centering
\includegraphics[width=0.65\linewidth]{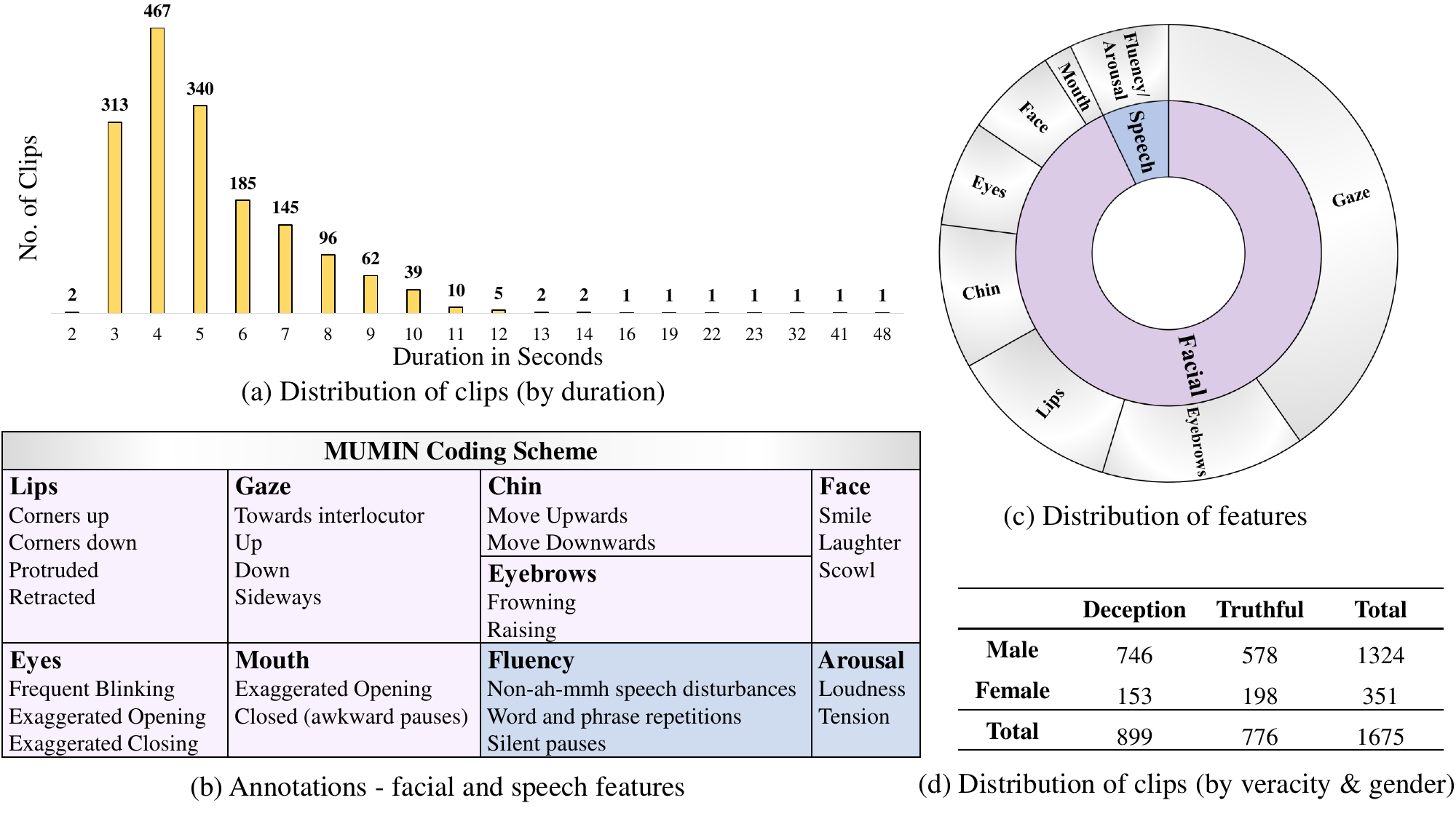}
  \caption{\small{DOLOS dataset for multimodal deception detection. Zoom in for a better view.}
  }
\label{fig:distribution}
\end{figure*}   

\section{The DOLOS Dataset}
\vspace{0.2em}
\label{sec:formatting}

\subsection{Gameshow Format} 

We collect data from a British reality comedy gameshow available on YouTube~\footnote{Videos are collected from YouTube under the fair use policy}, where six participants compete as two teams. In each taking turn, one of their members read out a statement about their personal life from a card. This statement could be either true or false, and only the speaker knows which one it is. The opposing team then asks the speaker a series of questions, and the speaker must provide answers in defense of their statement. The goal of the speaker is to convince the opposing team that their statement is true, while the opposing team tries to determine whether the statement is actually true or false based solely on the speaker's answers. At the end of each round, the veracity of the statement is revealed, and if the opposing team correctly predicts whether it was true or false, they win the round; otherwise, the speaker's team wins. The game continues for several rounds until all participants have played. Notably, if the statement is true, then the speaker's answers will be considered as truthful video clips and vice versa. The gameshow format is given in Fig.~\ref{fig:format}. 

\subsection{Data Collection and Annotation Procedure \label{DOLOS anno method}}

For each episode of the gameshow videos, the video clips that satisfy the following requirements are extracted:
\vspace{-0.5em}
\begin{itemize}
  \item The participant speaks only the relevant content (\emph{i.e.}, telling the truth or lies) in a clear voice and without strong background noise.
  \vspace{-0.5em}
  \item The face of the participant is clearly visible without occlusion.
\vspace{-0.5em}
\end{itemize}

\newcommand{\tabincell}[2]{\begin{tabular}{@{}#1@{}}#2\end{tabular}}
\begin{table*}[]
\centering
\label{table:dataset comparison table}
\begin{adjustbox}{max width=0.8\textwidth}
    \begin{tabular}{cccccccc}
    \toprule
    \multirow{2}{*}{\textbf{Dataset}} & \multirow{2}{*}{\tabincell{c}{\textbf{Hand annotated} \\  \textbf{features}} }& \multirow{2}{*}{\textbf{\#Subjects}} & \multicolumn{4}{c}{\textbf{Samples}} & \multirow{2}{*}{\textbf{Scenario}} \\ %\cline{4-7}
     &  &  & \multicolumn{1}{c}{\textbf{Total}} & \multicolumn{1}{c}{\textbf{Deceptive}} & \multicolumn{1}{c}{\textbf{Truthful}} & \textbf{Deception/Truth Ratio} &  \\
     \midrule
    Real Life Trials\cite{perez2015deception} & \checkmark & 56 & \multicolumn{1}{c}{121} & \multicolumn{1}{c}{61} & \multicolumn{1}{c}{60} & 1.02 & Court-room \\
    Real Life Deception\cite{perez2015verbal} &  & - & \multicolumn{1}{c}{56} & \multicolumn{1}{c}{31} & \multicolumn{1}{c}{25} & 1.24 & Street interviews \\
    Political Deception\cite{kamboj2020multimodal} &  & 88 & \multicolumn{1}{c}{180} & \multicolumn{1}{c}{-} & \multicolumn{1}{c}{-} & - & Political speech \\
    \midrule
    Bag of Lies\cite{gupta2019bag} &  & 35 & \multicolumn{1}{c}{325} & \multicolumn{1}{c}{162} & \multicolumn{1}{c}{163} & 0.99 & Lab \\
    MU3D\cite{lloyd2019miami} &  & 80 & \multicolumn{1}{c}{320} & \multicolumn{1}{c}{160} & \multicolumn{1}{c}{160} & 1 & Lab \\
    Deception Detection and remote PPG\cite{speth2021deception} &  & 70 & \multicolumn{1}{c}{1680} & \multicolumn{1}{c}{630} & \multicolumn{1}{c}{1050} & 0.6 & Lab \\
    \midrule
    Box of Lies\cite{soldner2019box} & \checkmark & 26 & \multicolumn{1}{c}{1049} & \multicolumn{1}{c}{862} & \multicolumn{1}{c}{187} & 4.61 & Gameshows \\
    \rowcolor{Gainsboro!60} \textbf{DOLOS (Ours)} & \checkmark & 213 & \multicolumn{1}{c}{1675} & \multicolumn{1}{c}{899} & \multicolumn{1}{c}{776} & 1.16 & Gameshows \\
    \bottomrule
    \end{tabular}
\end{adjustbox}
\caption{Comparison of multimodal deception detection datasets}
\vspace{-1.5em}
\label{table: dataset comparison table}
\end{table*}

From a total of 84 episodes, we extracted 1675 video clips from 213 (141 Male and 72 Female) participants. The duration of the video clips ranges from 2-19 seconds. The extracted video clips are manually annotated for non-verbal deceptive cues using the popular MUMIN coding scheme \cite{allwood2005mumin}, where we particularly focus on the visual (25 facial) and vocal (5 speech) features. The non-verbal features in the MUMIN coding scheme and the distributions of video clips are shown in Fig.~\ref{fig:distribution}. 

To eliminate bias among the human annotators,  all six human annotators underwent calibration by annotating the same subset of the dataset, and their performance was evaluated using Cohen's Kappa scores~\cite{banerjee1999beyond}. If significant discrepancies are found in visual and audio annotations between annotators, they reconciled their differences through discussion and then re-annotated the subset. This iterative process was repeated until inter-annotator agreement was achieved. Before calibration, the average Cohen's Kappa scores were 0.5 (0.53 for audio and 0.47 for visual annotations). After calibration, the average scores increased to 0.65 (0.67 for audio and 0.63 for visual annotations).

Based on the statistics in Fig.~\ref{fig:distribution}, we organize DOLOS dataset into three different protocols. First, we provide a train-test protocol with a 3-fold split to evaluate the performance of the deception detector for DOLOS dataset. Second, to reflect the variability in speaking duration in real-world scenarios, we provide a duration protocol with short clips (2-4s) and long clips (5-10s) based on the statistics from our dataset. We also provide a gender (male and female) protocol to investigate different factors in the deception detection task.

\subsection{Comparison with Box of Lies and Other Datasets}

We compare our DOLOS with Box of Lies (BOL) \cite{soldner2019box}, which was also collected from gameshow videos. In terms of quantity, BOL has fewer video clips and fewer subjects. BOL contains only 26 subjects, which is 17.9\% of DOLOS. In terms of quality, among 1,049 video clips in BOL, 537 video clips missed the speaker's face. In the other video clips, the speaker and the audio of some video clips do not match. In BOL, the host of the gameshow has more video clips than other subjects, which introduces a large bias. In comparison, the data collection requirements mentioned in Sec.~\ref{DOLOS anno method} inherently prevent these problems. Therefore, DOLOS is more suitable for audio-visual deception detection research (see Sec.~\ref{cross-testing}) than BOL.

We also compare DOLOS with other publicly available datasets. A summary of the comparisons is shown in Table \ref{table: dataset comparison table}. Briefly, DOLOS offers the following advantages:
\begin{itemize}
\vspace{-0.5em}
\item DOLOS is the largest in terms of number of subjects. It is also the largest in terms of video clips in the non-lab scenario.
\vspace{-0.5em}
\item Compared to the Box of Lies gameshow dataset, DOLOS contains a larger number of video clips and is more well-balanced in terms of the proportion of deceptive and truthful samples.
\vspace{-0.5em}
\item DOLOS provides manually annotated MUMIN features for all video clips. These annotations can be directly used for deception detection task. They can also be applied for other related tasks (\emph{e.g.}, facial expression prediction) or multi-task learning. 
\vspace{-0.5em}
\end{itemize}

\begin{figure*}[t]
\centering
\includegraphics[width=0.6\linewidth]{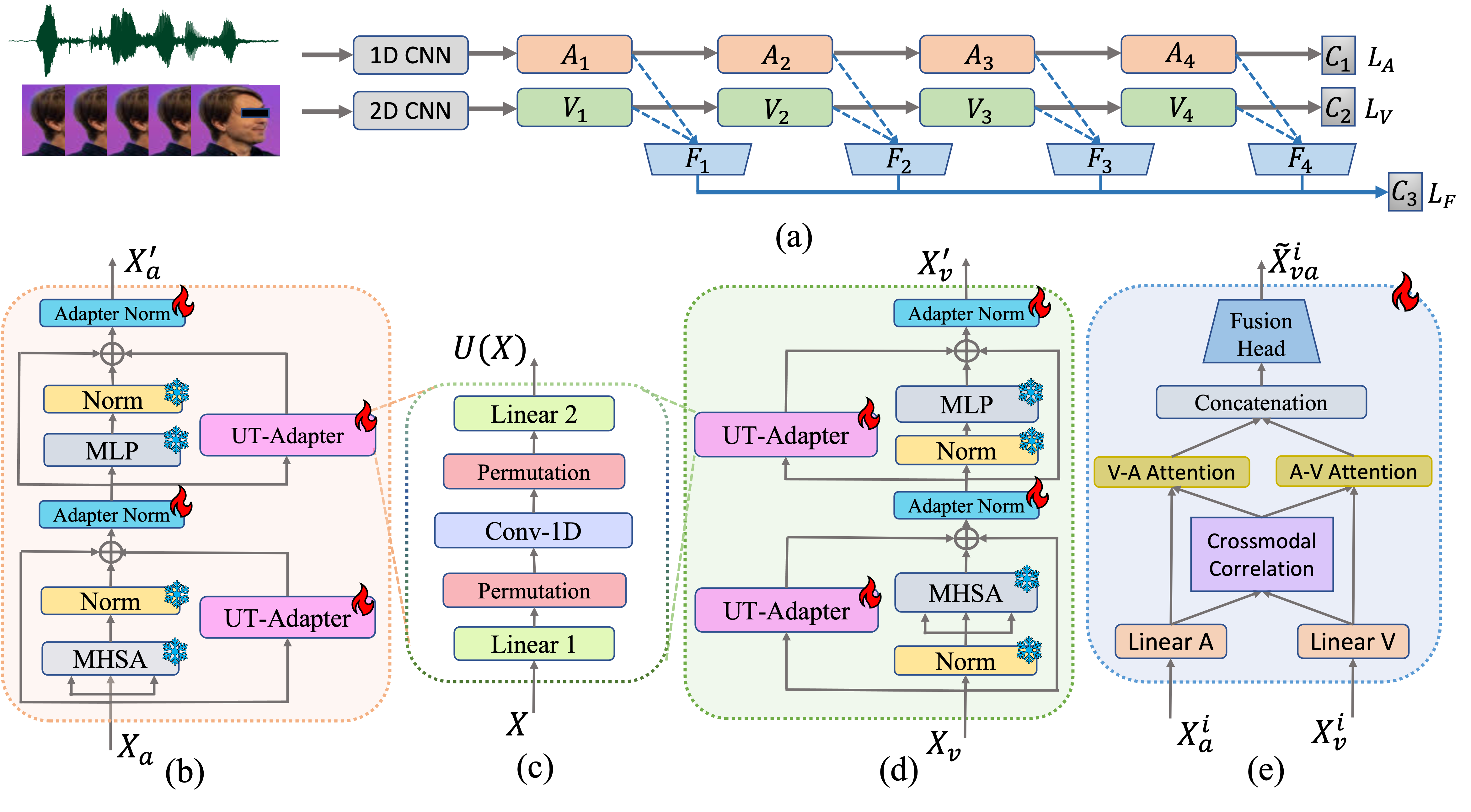}
  \caption{\small{Parameter-Efficient Crossmodal Learning (PECL) Framework. (a) is the main architecture. (b) is the encoder layer $A_i$ for audio features. (c) is the UT-Adapter with a shared structure between audio and visual features. (d) is the encoder layer $V_i$ for visual features. (e) is the PAVF module $F_i$. The trainable modules and layers are marked by ``fire", and the frozen layers are marked by ``snowflakes" .}}
\label{fig:main}
\vspace{-1.0em}
\end{figure*}

\section{Methodology}
\vspace{0.2em}

To efficiently leverage the prior knowledge from large-scale generic audio and visual tasks, we propose Parameter-Efficient Crossmodal Learning (PECL) for audio-visual deception detection. PECL's overall framework is illustrated in Fig.~\ref{fig:main} with the main network architecture (Fig.~\ref{fig:main} (a)) and several components with detailed structures (Fig.~\ref{fig:main} (b)-(e)).

Specifically, in Fig.~\ref{fig:main} (a), visual inputs are tokenized by a 2D-CNN module and audio inputs are tokenized by a 1D-CNN module (See Sec.~\ref{more details}) with the same dimensions, which are denoted as $\boldsymbol{X_v} \in \mathbb{R} ^ {L \times D}$ and $\boldsymbol{X_a} \in \mathbb{R} ^ {L \times D}$, where $L$ is the length of sequences and $D$ is the dimension of each sequence embedding. $V_i$ and $A_i$ represent transformer-based visual and audio modality encoder layers, where we adopt W2V2~\cite{baevski2020wav2vec} as the backbone network for audio modality and ViT~\cite{dosovitskiy2020vit} for visual modality. $F_i$ are fusion modules between each pair of unimodal encoder layers and $i={1, 2, 3, 4}$. $C_1$, $C_2$, and $C_3$ are classifiers for audio, visual, and fusion networks, respectively. $L_A$, $L_V$, and $L_F$ are cross-entropy losses for audio, visual, and fused modalities, respectively. 

\subsection{Uniform Temporal Adapter (UT-Adapter) \label{UT-adapter}}

To improve the parameter efficiency and alleviate the overfitting issue, we insert learnable adapter layers~\cite{houlsby2019parameter, jie2022convolutional} in the pre-trained visual and audio models. During training, all the weights, except those in the adapter layers and the classification layers are frozen. Unlike NLP-Adapter for language tasks~\cite{houlsby2019parameter} and Conv-Adapter for vision tasks~\cite{jie2022convolutional}, our adapter layer, named Uniform Temporal Adapter (UT-Adapter), is to explore temporal attention for both visual and audio modalities.

Let $F_V$ and $G_A$ represent the visual and audio encoder layers in Fig.~\ref{fig:main}. Their outputs are denoted as
\vspace{-0.5em}
\begin{equation}\small
\label{eq1}
    \boldsymbol{X^{'}_v} = F_V (\boldsymbol{X_v})\ and \
    \boldsymbol{X^{'}_a} = G_A (\boldsymbol{X_a}),
\end{equation}
\noindent where $\boldsymbol{X^{'}_v} \in \mathbb{R} ^ {L \times D}$ and $\boldsymbol{X^{'}_a} \in \mathbb{R} ^ {L \times D}$. $F_V$ and $G_A$ are formed by a Multi-Head Self-Attention (MHSA) module, UT-Adapters, a Multi-Layer Perception ($MLP$) module, and LayerNorm ($LN$). To simplify the notations, let $\boldsymbol{X}$ be one of the input features, \emph{i.e}, $\boldsymbol{X_v}$ and $\boldsymbol{X_a}$ (in Fig.~\ref{fig:main} (b) and (d)). $\boldsymbol{X}$ is projected to a query $\boldsymbol{Q} = \boldsymbol{XW_Q}$, a key $\boldsymbol{K} = \boldsymbol{XW_K}$, and a value $\boldsymbol{V} = \boldsymbol{XW_V}$, where $\boldsymbol{W_Q}$, $\boldsymbol{W_K}$, and $\boldsymbol{W_V} \in \mathbb{R}^ {D \times D}$ are pre-trained projection weights. For a single head $H_j, j \in \{1, 2, \cdots, N\}$, the self-attention operation is formulated as
\vspace{-0.2em}
\begin{equation}\small
\label{eq2}
    H_j (\boldsymbol{X}) = Softmax \left(\cfrac{\boldsymbol{Q_j}\boldsymbol{K_j}^{\intercal}}{\sqrt{D}}\right)\boldsymbol{V_j},
\end{equation}
\noindent and the outputs of all the $H_j, j \in \{1, 2, \cdots, N\}$ are then concatenated and projected to a single output $H(\boldsymbol{X})$. 

UT-Adapters are placed in parallel with MHSA and MLP modules. In particular, a UT-Adapter is a stack of linear and 1D-convolutional layers. As shown in Fig.~\ref{fig:main} (c), the output of a UT-Adapter is operated as
\begin{equation}
\label{eq3}
    U(\boldsymbol{X}) = L_2(P(C(P(L_1(\boldsymbol{X};\boldsymbol{W_1}));\boldsymbol{W_C})); \boldsymbol{W_2}),
\end{equation}
\noindent where $L_1$ and $L_2$ are \emph{Linear 1} and \emph{Linear 2} layers in Fig.~\ref{fig:main} (c) and $P$ and $C$ are \emph{Permutation} and \emph{Conv-1D}, respectively. $\boldsymbol{W_1} \in \mathbb{R} ^ {D \times 128}$ and $\boldsymbol{W_2} \in \mathbb{R} ^ {128 \times D}$ are trainable weights of $L_1$ and $L_2$. $\boldsymbol{W_C}$ is the 1D-convolutional weight with the kernel size of 3. $L_1$ projects the input $\boldsymbol{X} \in \mathbb{R} ^ {L \times D}$ to $ \boldsymbol{X} \in \mathbb{R} ^ {L \times 128}$, Permutation layer $P$ shifts $ \boldsymbol{X} \in \mathbb{R} ^ {L \times 128}$ to $ \boldsymbol{X} \in \mathbb{R} ^ {128 \times L}$ and $C$ is applied along the temporal dimension to capture the temporal dynamics. $P$ layer and $L_2$ then project $ \boldsymbol{X} \in \mathbb{R} ^ {128 \times L}$ to $ \boldsymbol{X} \in \mathbb{R} ^ {L \times D}$.

To be specific, for $F_V$, the output from MHSA and a UT-adapter can be represented as:
\begin{equation}\small
\label{eq4}
    \boldsymbol{X^{''}_v} = AN(\boldsymbol{X_v} + H(LN(\boldsymbol{X_v})) + U(\boldsymbol{X_v})),
\end{equation}
\noindent and $\boldsymbol{X^{'}_v}$ in Eq.~\ref{eq1} is obtained by:
\begin{equation}\small
\label{eq5}
    \boldsymbol{X^{'}_v} = AN(\boldsymbol{X^{''}_v} + MLP(LN(\boldsymbol{X^{''}_v})) + U(\boldsymbol{X^{''}_v})),
\end{equation}
\noindent where $AN$ is a normalization layer introduced after UT-Adapter, which is trainable.

$G_A$ shares a similar structure with $F_V$. The output $\boldsymbol{X_a}$ is computed by
\vspace{-0.3em}
\begin{equation}\small
\label{eq6}
\begin{split}
    \boldsymbol{X^{''}_a} = AN(\boldsymbol{X_a} &+ LN(H(\boldsymbol{X_a})) + U(\boldsymbol{X_a})), \\
    \boldsymbol{X^{'}_a} = AN(\boldsymbol{X^{''}_a} &+ LN(MLP(\boldsymbol{X^{''}_a})) + U(\boldsymbol{X^{''}_a})). 
\end{split}
\end{equation}

The input $\boldsymbol{X_v}$ and $\boldsymbol{X_a}$ are sequence embeddings from vision and audio. The MHSA module in the transformer encoder facilitates global attention, allowing the model to effectively learn both spatial and temporal attention from the input. However, this may not be optimal for learning local temporal attention and local spatial information. Thus, UT-Adapter is proposed to capture the local temporal attention in parallel with MHSA and MLP modules. The 2D-CNN captures the local spatial information and the 1D-CNN captures temporal information. We empirically show that this \textit{adapter-transformer} architecture is parameter-efficient tuning and delivers better performance.

\subsection{Plug-in Audio-Visual Fusion (PAVF) \label{PAVF}}

To explore the audio-visual interactions, we propose a Plug-in Audio-Visual Fusion (PAVF) module that learns crossmodal attention for fusion. 

We denote the output features from $V_i$ and $A_i$ as $\boldsymbol{X_v^i} \in \mathbb{R} ^ {L \times D}$ and $\boldsymbol{X_a^i} \in \mathbb{R} ^ {L \times D}$ and propose to learn crossmodal attention in an embedding space with lower dimension, which is more efficient and reduces computational costs. $\boldsymbol{X_v^i}$ and $\boldsymbol{X_a^i}$ are first projected to $\boldsymbol{X_v^i} \in \mathbb{R} ^ {L \times D^{'}}$ and $\boldsymbol{X_a^i} \in \mathbb{R} ^ {L \times D^{'}}$, $D^{'} < D$, by linear projection layers. Crossmodal correlation $\boldsymbol{P^i} \in \mathbb{R}^{L \times L}$ is learned to indicate the importance between visual sequences and audio sequences. A specific visual sequence may have a higher correlation to certain audio sequences in a video clip, which offers important cues to the deception detection task. To calculate $\boldsymbol{P^i}$, we introduce a trainable crossmodal correlation weight matrix $\boldsymbol{W_P^i \in \mathbb{R}^{D^{'} \times D^{'}}}$ and the calculation of $\boldsymbol{P^i}$ is
\vspace{-0.5em}
\begin{equation}
\label{eq7}
    \boldsymbol{P^i} = \boldsymbol{X_a^i}\boldsymbol{W_P^i}\boldsymbol{X_v^i}^{\intercal}.
\end{equation}
\noindent We further use the learned crossmodal correlation $\boldsymbol{P^i}$, to perform $\boldsymbol{X_v^i} \rightarrow \boldsymbol{X_a^i}$ and $\boldsymbol{X_a^i} \rightarrow \boldsymbol{X_v^i}$  attentions. The attended features are
\begin{equation}\small
\label{eq8}
    \begin{split}
        \boldsymbol{\Tilde{X}_v^i} = Softmax(\boldsymbol{P^i})\boldsymbol{X_v^i} + \boldsymbol{X_v^i}, \\
        \boldsymbol{\Tilde{X}_a^i} = Softmax(\boldsymbol{P^i}^{\intercal})\boldsymbol{X_a^i} + \boldsymbol{X_a^i},
    \end{split}
\end{equation}
\noindent where $\boldsymbol{P^i}^{\intercal}$ is the transpose of $\boldsymbol{P^i}$, $\boldsymbol{\Tilde{X}_v^i} \in \mathbb{R} ^ {L \times D^{'}}$, and $\boldsymbol{\Tilde{X}_a^i} \in \mathbb{R} ^ {L \times D^{'}}$. Softmax is conducted column-wise. The attended features are concatenated and inputted into a fusion head. The fusion head consists of a linear projection $L_p$ that further reduces the embedding dimension to $D''$, a normalization layer $LN$, and a non-linear activation $ReLU$. It can be denoted as
\begin{equation}\small
\label{eq9}
    \boldsymbol{\Tilde{X}_{va}^i} = ReLU(LN(L_p(\boldsymbol{\Tilde{X}_v^i} \oplus  \boldsymbol{\Tilde{X}_a^i}))),
\end{equation}
\noindent where $\boldsymbol{\Tilde{X}_{va}^i} \in \mathbb{R}^{L \times D^{''}}$ and $\oplus$ is a concatenation operation. 

PAVF modules are inserted into each pair of $V_i$ and $A_i$ to perform fusion, in which richer fused information can be learned. The outputs $\boldsymbol{\Tilde{X}_{va}^i}$ from each PAVF module are then concatenated and used for prediction.

\subsection{Multi-task Learning \label{multitask}}

Multi-task learning has demonstrated promising performance in various audiovisual tasks~\cite{hu2021unit,tao2020end}, effectively enhancing the performance of multiple tasks through mutual learning. With DOLOS' MUMIN features (Fig.~\ref{fig:distribution} (b)), we enhance performance through multi-task learning, simultaneously conducting two prediction tasks using fused multimodal features. More clearly, for $K$ MUMIN features labels and one deception-truth label, we predict $K+1$ labels. Note that all the labels are binary. We denote the fused audio-visual embedding as $\boldsymbol{\Tilde{X}_{va}}$. The classification layer takes $\boldsymbol{\Tilde{X}_{va}}$ as input to produce prediction scores $S \in \mathbb{R}^{1 \times (K+1)}$. Given the ground-truth label $Y \in \mathbb{R}^{1 \times (K+1)}$, we use cross-entropy loss to perform multi-task learning. It can be formulated as
\begin{equation}\small
\label{eq10}
    \mathcal{L}_M = - \sum_{k=1}^{K+1}\left( Y_k\log(S_k) + (1 - Y_k)\log(1 - S_k) \right),
\end{equation}
\noindent where $\mathcal{L}_M$ is the sum of the cross-entropy losses for each element in $S$ and $Y$.

% Intra-Testing table
\begin{table*}[]
\centering
\resizebox{0.6\textwidth}{!}{%
\begin{tabular}{cccc|ccc|ccc}
    \toprule
    \multirow{2}{*}{\textbf{Modalities}} & \multicolumn{3}{c}{\textbf{3-Fold Average}} & \multicolumn{3}{c}{\textbf{Duration Protocol}} & \multicolumn{3}{c}{\textbf{Gender Protocol}} \\
     & ACC & F1 & AUC & ACC & F1 & AUC & ACC & F1 & AUC \\
    \midrule
    Visual & 61.44 & 69.42 & 58.89 & 61.03 & \textbf{72.01} & 56.51 & 59.37 & 64.19 & 54.94 \\
    Audio & 59.19 & \textbf{73.46} & 52.54 & 58.24 & 71.83 & 52.38 & 52.62 & 63.22 & 51.08 \\
    \midrule
    Concatenation & 61.62 & 70.2 & 60.5 & 60.8 & 69.63 & 57.79 & 58.0 & 66.22 & 53.75 \\
    \rowcolor{Gainsboro!60} Fusion (PAVF) & 64.75 & 71.2 & 62.71 & 62.43 & 70.04 & 59.92 & 58.28 & 65.41 & 53.31 \\
    \rowcolor{Gainsboro!60} PAVF + Multi-task & \textbf{66.84} & 73.35 & \textbf{64.58} & \textbf{64.48} & 71.09 & \textbf{62.44} & \textbf{59.04} & \textbf{66.84} & \textbf{55.1} \\
    \bottomrule
\end{tabular}
}
\caption{Multimodal deception detection on DOLOS dataset. The metrics are ACC (\%), F1 (\%), and AUC(\%).}
\label{table:intra-testing}
% \vspace{-1.0em}
\end{table*}

% Cross-Testing table
\begin{table}[]
\centering
\begin{adjustbox}{max width=0.5\textwidth}
\begin{tabular}{cccc|cc}
    \toprule
     &\multirow{2}{*}{\textbf{Method}}& \multicolumn{2}{c}{\textbf{Train Data - DOLOS}} & \multicolumn{2}{c}{\textbf{Train Data - Box of Lies}}  \\
     & &Test BgOL & Test RLT & Test BgOL & Test RLT \\
    \midrule
    \multirow{3}{*}{V} &  RN18+LSTM  &  55.08 (+4.62) & 56.62 (+3.85) & 50.46 & 52.77 \\
     &3D RN  & 55.69 (+3.38) & \textbf{57.87} (+\textbf{5.53}) & 52.31 & 52.34 \\
    &Ours (Vision) & \textbf{56.92} (+4.30) & 55.32 (+2.55) & 52.62 & 52.77 \\
    \midrule
    \multirow{3}{*}{A} & RN18  $\dag$&  \textbf{55.84} (+\textbf{4.91}) & 53.53 (+1.64) & 50.93 & 51.89 \\
    &MLP (MFCC)  &  52.31 (+2.46) & 52.34 (+0.42) & 49.85 &  51.92\\
    & Ours (Audio) $\ddag$ &55.39 (+4.00) & \textbf{59.15} (+\textbf{4.68}) & 51.39 & 54.47 \\
    \midrule
    \multirow{3}{*}{V+A} & 3D RN+RN18 & 50.77 (+1.23) & 54.47 (+1.70) & 49.54 & 52.77 \\
    % &  &  (+) &  (+) &  &  \\
     &  Ours (PAVF) &  \textbf{57.54} (+\textbf{4.31}) &  \textbf{56.17} (+\textbf{2.98}) &  53.23 & 53.19 \\
    \bottomrule
\end{tabular}
\end{adjustbox}
\caption{Cross-testing on Bag-of-Lies (BgOL) and  Real-Life-trials (RLT) datasets. The metric is ACC (\%). RN stands for ResNet. All the visual features are from face frames. $\dag$ is trained on mel-spectrogram and $\ddag$ is trained on raw audio.}
\vspace{-0.8em}
\label{table:cross-testing}
\end{table}

\section{Experimental Results}

\subsection{Implementation Details \label{implementation details}}

\noindent\textbf{Data pre-processing.} For each video clip, we evenly sampled $L=64$ images and cropped the face areas using MTCNN face detector~\cite{zhang2016joint}. These images were normalized and resized to $160 \times 160$ pixels. The raw speech audio was resampled such that the W2V2 feature extractor outputs $L=64$ tokens. No data augmentation was applied to both modalities.

\noindent\textbf{Model details.\label{more details}} We used ImageNet pre-trained ViT as the backbone network for visual modality. We tokenized face images with a 2D-CNN module, which resulted in a feature with a dimension of $64 \times 256$. For the audio modality, we adopted the pre-trained W2V2 model. The raw audio was tokenized by the 1D-CNN module and the feature size was $64 \times 512$ for each audio sample. Using a linear projection layer, visual and audio tokens were projected to $64 \times 768$ dimensions. We empirically utilized the first 4 transformer encoder layers from ViT and W2V2 models to extract features because deception detection can benefit from the low-level features~\cite{stathopoulos2020deception,krishnamurthy2018deep,gogate2017deep}. The UT-Adapter and PAVF modules were inserted in these four encoder layers (in Fig.~\ref{fig:main} (a)). The one-dimensional convolution layer (in Fig~\ref{fig:main} (c)) in the UT-Adapter for both visual and audio encoders had a kernel size of 3 and a stride of 1. 

\noindent\textbf{Training and evaluation.} The models were trained with cross-entropy loss. We used Adam optimizer and trained the models for 20 epochs with a learning rate of 3e-4 and a batch size of 16. We evaluated our method with accuracy (ACC), F1 score (F1), and Area Under Curve (AUC). 

\subsection{Audio-visual Deception Detection in DOLOS}

The performance of PECL model on DOLOS is presented in Table~\ref{table:intra-testing}. The average results were calculated from the 3 folds defined in the train-test protocol. Duration and gender protocol results were cross-tested averages of training/testing on (long/short + short/long) and (male/female + female/male), respectively. For unimodal results, the visual modality showed slightly better performance than the audio across all the protocols. To study the performance of a simple feature-level fusion, we concatenated the last output embeddings from visual and audio encoders to perform prediction. The concatenation usually draws the average of unimodal predictors.
To overcome this issue, the proposed PAVF module learned the correlation between visual and audio features from different encoder layers and improved the performance. The multi-task learning further boosted the accuracy across all the protocols. However, by comparing the overall performances, we observed that the duration and gender factors showed an adverse effect on performance. These results pointed out the importance of duration and gender factors in deception detection.

\begin{table}[t]
\centering
\begin{adjustbox}{max width=0.48\textwidth}
\begin{tabular}{cccc}
    \toprule
     & \textbf{Features} & \textbf{Method} & \textbf{ACC (\%)} \\
     \midrule
    \multirow{5}{*}{V} & Open Face~\cite{mathur2020introducing,krishnamurthy2018deep} & LSTM & 56.81 \\
     & Action Units (AU)~\cite{mathur2020introducing,avola2019automatic} & LSTM & 57.47 \\
     & Facial Affect~\cite{mathur2020introducing,karimi2018toward} & LSTM & 57.67 \\
     & MUMIN Features~\cite{perez2015deception,perez2015verbal} & MLP & 58.84 \\
     & Face ~\cite{karnati2021lienet,ding2019face,wu2018deception,gogate2017deep,krishnamurthy2018deep,karimi2018toward} & RN18+LSTM & 59.55 \\
      & \cellcolor{Gainsboro!60} Face &\cellcolor{Gainsboro!60} Ours (Vision) &\cellcolor{Gainsboro!60} \textbf{61.44} \\ \hline
    \multirow{3}{*}{A} & MFCC~\cite{wu2018deception} & MLP & 56.86 \\
     & Open SMILE~\cite{mathur2020introducing,gogate2017deep,yang2021multimodal} & MLP & 57.49 \\
     & \cellcolor{Gainsboro!60} W2V2 Audio~\cite{karnati2021lienet,ding2019face,krishnamurthy2018deep,karimi2018toward} & \cellcolor{Gainsboro!60}Ours (Audio) & \cellcolor{Gainsboro!60} \textbf{59.19} \\ \hline
     & Open Face + Open SMILE~\cite{mathur2020introducing} & Score-Fusion & 59.75 \\
    \multirow{5}{*}{V+A} & Face + Open SMILE~\cite{gogate2017deep,krishnamurthy2018deep,yang2021multimodal} & Score-Fusion & 60.00 \\
     & \cellcolor{Gainsboro!60} Face + W2V2 Audio~\cite{karnati2021lienet,karimi2018toward,ding2019face} & \cellcolor{Gainsboro!60}Ours(Concat) &\cellcolor{Gainsboro!60} \textbf{61.62} \\
     & \cellcolor{Gainsboro!60} Face + W2V2 Audio &\cellcolor{Gainsboro!60} Ours(PAVF) &\cellcolor{Gainsboro!60} \textbf{64.75}\\
    \bottomrule
\end{tabular}%
\end{adjustbox}
\caption{Benchmarking DOLOS dataset on visual and audio features. The weighted average is used for score fusion. RN18 stands for ResNet-18.}
\label{table:benchmark}
\vspace{-1.3em}
\end{table}

\vspace{-0.3em}
\subsection{Comparison with Box of Lies\label{cross-testing}}
\vspace{-0.2em}

To compare DOLOS with the current gameshow dataset BOL, we conducted cross-testing, where DOLOS and BOL were used for training and other deception detection datasets, \emph{i.e.}, Bag of Lies and Real Life Trials, for testing. It was challenging because these datasets were collected from different deceptive scenarios. As shown in Table~\ref{table:cross-testing}, we compared several methods for audio, visual, and fusion and found that models trained on DOLOS achieved higher accuracies overall. PECL model trained on DOLOS also showed better test accuracy than that trained on BOL, (\emph{e.g.}, Fusion (PAVF) improved 4.31\% on BgOL and 2.98\% on RLT). These results suggest that models trained on DOLOS can be more effective in identifying deceptive content in other scenarios than those trained on BOL.

\vspace{-0.4em}
\subsection{Benchmarking DOLOS}
\vspace{-0.4em}
We benchmark DOLOS in Table~\ref{table:benchmark}, following the deception detection literature. It is worth noting that the existing literature on deception detection is not standardized with protocols, data processing, and feature extraction methodologies. To highlight and cover the most representative works in deception detection literature, we benchmarked DOLOS on the popular visual and audio features used before for deception detection. The OpenFace features (2D facial landmarks, eye gaze, and head pose) and Action Units (AU) features were extracted using the OpenFace toolkit~\cite{amos2016openface}. The facial affect (emotion) features were extracted using the Affectnet~\cite{mollahosseini2017affectnet} model. For audio, the Mel Cepstral Frequency Coefficients (MFCC) and openSMILE features were extracted by the openSMILE toolkit~\cite{eyben2010opensmile}.

\vspace{0.3em}
\noindent\textbf{Performance.} For visual modality, the RGB face images gave the best performance, since they contained more deceptive cues. This was followed by manually annotated MUMIN features, affect features, and AU features, which all represented high-level human expressions. The low-level facial landmark features did not perform well. For audio modality, raw speech audio with rich deceptive cues performed the best. The openSMILE features performed better than MFCC, since it extracted nearly three times the audio features compared with MFCC. Score-level fusion was conducted for the best-performing features of the two modalities. The proposed transformer-based detector achieved the best in both unimodal and multimodal settings.

\subsection{Ablation Study}
\vspace{-0.5em}
The ablation studies were conducted on different components of PECL model based on the train-test protocol. 
% ablation figure
\begin{figure}[t]
\centering
\includegraphics[width=0.65\linewidth]{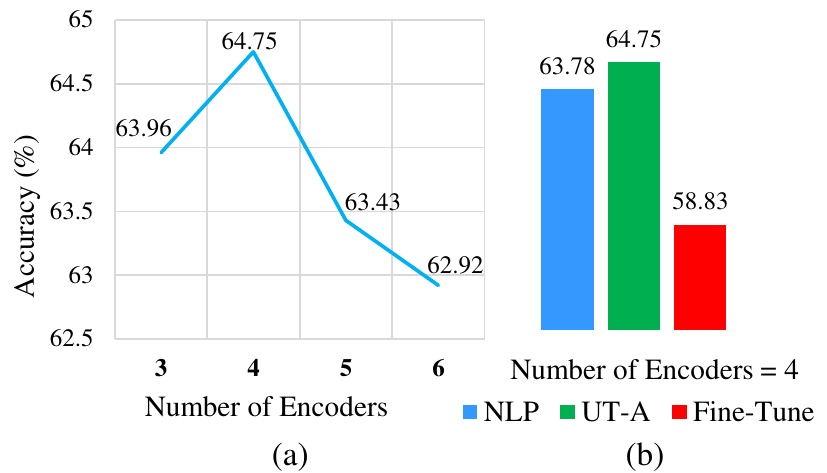}
  \caption{\small{ Ablation on (a) encoder depth and (b) type of adapter.}}
\label{fig:ablation}
\vspace{-0.8em}
\end{figure}

\begin{table}[t]
\centering
\begin{adjustbox}{width=\columnwidth,center}
\scalebox{0.2}{
\begin{tabular}{c|c|c|c||c|c}
\toprule
 \multirow{2}{*}{\textbf{Method}} &  \multicolumn{3}{c}{\textbf{Trainable}} & \multirow{2}{*}{\textbf{Total}} & \multirow{2}{*}{\textbf{Trainable/Total}}\\ 
   & Adapter & PAVF & Total & &\\ 
\midrule
  Fine-Tune & $-$  & 1.982 & 70.180 & 70.180 & 100\% \\
 NLP~\cite{houlsby2019parameter} & 5.345 &1.982 &7.327  &73.347 & 9.99\%\\ 
 UT-A & 3.077 & 1.982 & \textbf{5.059} & 71.079 & \textbf{7.12\%}\\ 

 \bottomrule
\end{tabular}
}
\end{adjustbox}
\caption{Comparison of the number of trainable parameters and total parameters (Millions).}
\label{table:params numbers}
\end{table}

\begin{table}[t]
\centering
\begin{adjustbox}{width=\columnwidth,center}
\scalebox{0.2}{
\begin{tabular}{c|c|c|c|c}
\toprule
Position & $\shortparallel$ MHSA $\shortparallel$ FFN & $\shortparallel$ MHSA & $\shortparallel$ FFN & MHSA $\triangle$ FFN\\ 
\midrule
 ACC (\%) & \textbf{64.75} &   63.88   &   63.37 &  64.39 \\
 \bottomrule
\end{tabular}
}
\end{adjustbox}
\caption{Ablation study on UT-Adapter positions. $\shortparallel$ indicates ``in parallel with" and $\triangle$ indicates ``between".}
\vspace{-0.8em}
\label{table:adapter position ablation}
\end{table}

\vspace{0.3em}
\noindent\textbf{Impact of encoder depth and adapter types.} Fig.~\ref{fig:ablation} (a) indicated that the proposed PECL performed best with four layers and may overfit when additional layers were added. Fig.~\ref{fig:ablation} (b) showed that UT-Adapter outperformed the NLP adapter~\cite{houlsby2019parameter} and full fine-tuning, which may be due to better temporal modeling with Conv1D layers in UT-Adapter.

\vspace{0.3em}
\noindent\textbf{Comparison of parameter amount.} Table~\ref{table:params numbers} demonstrated that our proposed method achieved the best performance while utilizing the least amount of trainable parameters when compared to both the NLP adapter and full fine-tuning approaches. Note that UT-Adapter performed best with 128 dimensions while NLP adapter with 256 dimensions.

\vspace{0.3em}
\noindent\textbf{Impact of adapter position.} The results in Table \ref{table:adapter position ablation} showed that the UT-Adapter worked the best when parallel to both MHSA and FFN layers~\cite{jie2022convolutional}, as they both captured the temporal attention. The performance dropped when parallel to only either layer. The performance was also close for UT-Adapter placed in-between MHSA and FFN layers.
\vspace{0.5em}

\begin{figure}[t]
\centering
\includegraphics[width=0.55\linewidth]{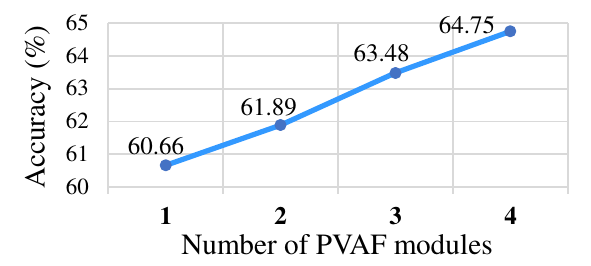}
% \vspace{-0.9em}
\caption{\small{Ablation on numbers of PAVF modules.}}
\label{fig:fusion_ablation}
\vspace{-0.8em}
\end{figure}

\begin{figure}[t]
\centering
\includegraphics[width=0.55\linewidth]{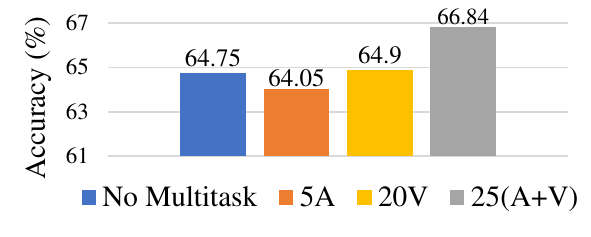}
% \vspace{-1.2em}
\caption{\small{Results of multi-task learning with different features.}}
\label{fig:multitask_ablation}
\vspace{-1.5em}
\end{figure}

\vspace{0.3em}
\noindent\textbf{Impact of PAVF modules.} The number of PAVF modules in Fig.~\ref{fig:fusion_ablation} corresponds to the last number of audio-visual encoders that operate with them. The advantage of fusion in earlier stages~\cite{nagrani2021attention} can be clearly seen with a substantial increase in performance for four PAVF modules, where the low-level crossmodal interactions were captured.
The results revealed that four PAVF modules are particularly helpful in capturing audio-visual deceptive cues.

\vspace{0.3em}
\noindent\textbf{Impact of multi-task learning.} In Fig.~\ref{fig:multitask_ablation}, we illustrated the ablation study of multi-task learning described in Sec~\ref{multitask}, for different audio-visual features (Refer Fig~\ref{fig:distribution} (b)). The results showed that using 20 visual features (20V) achieved better results than using 5 audio features (5A) and using 25 audio-visual features further boosted accuracy. The results demonstrated that the manually annotated features in DOLOS can benefit deception detection performance.

\vspace{-0.7em}
\section{Conclusion}
\vspace{-0.5em}
In this paper, we introduced DOLOS, the largest gameshow deception dataset with fine-grained audio-visual feature annotations. We also proposed Parameter-Efficient Crossmodal Learning, where the UT-Adapter learns temporal attention for both visual and audio modalities, and the PAVF module captures correlation information between audio and visual modalities. We benchmarked our dataset on the existing deception detection approaches. Extensive experiments showed that DOLOS has better quality than the Box of Lies and PECL model achieved better results on DOLOS and can be applied to other deceptive content. 
%In the future, we will also investigate language modality and other open issues such as domain generalization in deception detection. We hope our work can inspire the community and attract more attention to the deception detection task.

\textbf{Ethical Consideration.}\quad Deception detection research is crucial in credibility assessment, anti-frauds, border security, etc. However, the misuse of deception detection may cause potential negative impacts such as privacy invasion, discrimination, and/or harmful surveillance. This work was completed under an ethics clearance (IRB-2022-901). Data was collected from YouTube under the fair use policy and the subjects are public personalities.

\textbf{Acknowledgments.}\quad This work was done at Rapid-Rich Object Search (ROSE) Lab, EEE, Nanyang Technological University. This research is supported in part by the NTU-PKU Joint Research Institute (a collaboration between the Nanyang Technological University and Peking University that is sponsored by a donation from the Ng Teng Fong Charitable Foundation), and the DSO National Laboratories, Singapore, under the project agreement No. DSOCL21238.

%-------------------------------------------------------------------------

%%%%%%%%% REFERENCES
{\small
\bibliographystyle{ieee_fullname}
\bibliography{egbib}
}

\end{document}